\documentclass[10pt,twocolumn,letterpaper]{article}

\usepackage[pagenumbers]{cvpr} 

\usepackage{graphicx}
\usepackage{amsmath}
\usepackage{amssymb}
\usepackage{booktabs}

\usepackage[pagebackref,breaklinks,colorlinks]{hyperref}

\usepackage[capitalize]{cleveref}
\crefname{section}{Sec.}{Secs.}
\Crefname{section}{Section}{Sections}
\Crefname{table}{Table}{Tables}
\crefname{table}{Tab.}{Tabs.}

\def\cvprPaperID{769} 
\def\confName{CVPR}
\def\confYear{2023}

\begin{document}

\title{Taming Transformer for Emotion-Controllable Talking Face Generation}

\author{Ziqi Zhang\\
Xidian University\\
{\tt\small zqzh9116@gmail.com}
\and
Cheng Deng\\
Xidian University\\
{\tt\small chdeng.xd@gmail.com}
}
\maketitle
\begin{abstract}
Talking face generation is a novel and challenging generation task, aiming at synthesizing a vivid speaking-face video given a specific audio.
To fulfill  emotion-controllable talking face generation, current methods need to overcome two challenges:
One is how to effectively model the multimodal relationship related to the specific emotion, and the other is how to leverage this relationship to synthesize identity-preserving emotional videos.
In this paper, we propose a novel method to tackle the emotion-controllable talking face generation task discretely.
Specifically, we employ two pre-training strategies to disentangle audio into independent components and quantize videos into combinations of visual tokens.
Subsequently, we propose the emotion-anchor (EA) representation that integrates the emotional information into visual tokens. 
Finally, we introduce an autoregressive transformer to model the  global distribution of  the visual tokens under the given conditions and further predict the index sequence for synthesizing the manipulated videos.
We conduct experiments on the MEAD dataset that controls the emotion of  videos conditioned on multiple emotional audios.
Extensive experiments demonstrate the superiorities of our method both qualitatively and quantitatively.

\end{abstract}

\section{Introduction}
\label{sec:intro}

\begin{figure}[tb]
	\centering
	\includegraphics[width=1\linewidth]{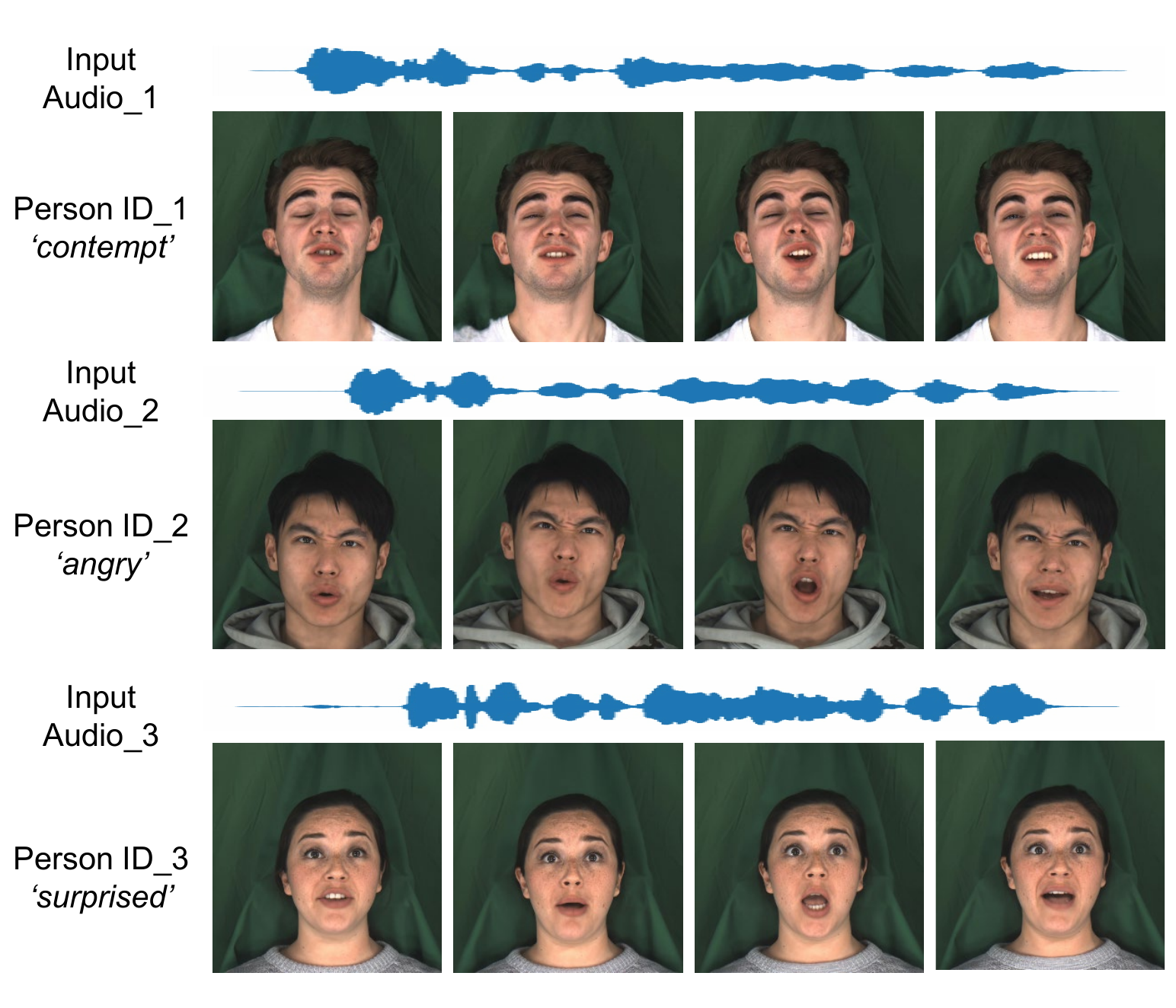}
	\caption{Illustration of emotion-controllable TFG.
		Given an extra audio clip and a target video, the emotion-controllable talking face generation method is capable of controlling the emotions (\textit{i.e., contempt, angry, surprised, etc.})  of the synthesized facial videos with identity-preservation. 
	}
	\label{intro}
\end{figure}

Talking Face Generation (TFG) aims to synthesize a specific talking face video conditioned both on the identity of the speaker (a frontal face photo) and the content of the speech (extra audio).
As one of the audio-driven cross-modality generation tasks, TFG attracts widespread interests in many multimedia applications, such as image animating \cite{liu2015video, edwards2016jali}, teleconferencing \cite{adalgeirsson2010mebot}, and enhancing speech comprehension \cite{eskimez2021speech}.
Current TFG methods mainly focus on the synchronizing of local actions and audio, such as lip synchronization or blinking.
Nevertheless, most previous methods miss a crucial part of  facial video generation, \textit{i.e.} the rich facial emotion of the speaker, which results in unnatural portraits.
To achieve emotion-controllable talking face generation shown in Fig. \ref{intro}, more challenges must be overcome, such as emotion representation and cross-modal modeling.
Although only some methods \cite{vougioukas2020realistic, chen2020talking} try to realize emotional control, they cannot fuse information extracted from different domains effectively and further model the relationship between emotion and facial generation.

Recently Wang \textit{et al.} \cite{wang2020mead} collected and released a Multi-view Emotional Audio-visual Dataset (MEAD), which contains several talking face videos with different emotions at different intensity levels.
To address the actual emotion rendering, they propose a method that sets one-hot vector as emotion conditions to control the generation.
However, this method renders only minor head movements without backgrounds, resulting in inconsistency over the face.
After that, Emotional Video Portraits (EVP)  \cite{ji2021audio} is proposed to focus on consistent emotion editing by learning to disentangle the content feature and emotion feature from the audio signals.
To model the relationship between the facial geometry and the emotion-induced local deformations within the face, Sinha \textit{et al.} \cite{sinha2022emotion} proposed a video editing method that leverages a graph convolutional neural network to learn facial geometry-aware landmark representation.
However, such methods rely on intermediate landmarks (or edge maps) to generate emotional textures, which are not conducive for networks to model the composition of images under different emotions.
So, they fail to synthesize the specific visual content that expresses the emotion of the audio  adequately.

\begin{figure}[tb]
	\centering
	\includegraphics[width=1\linewidth]{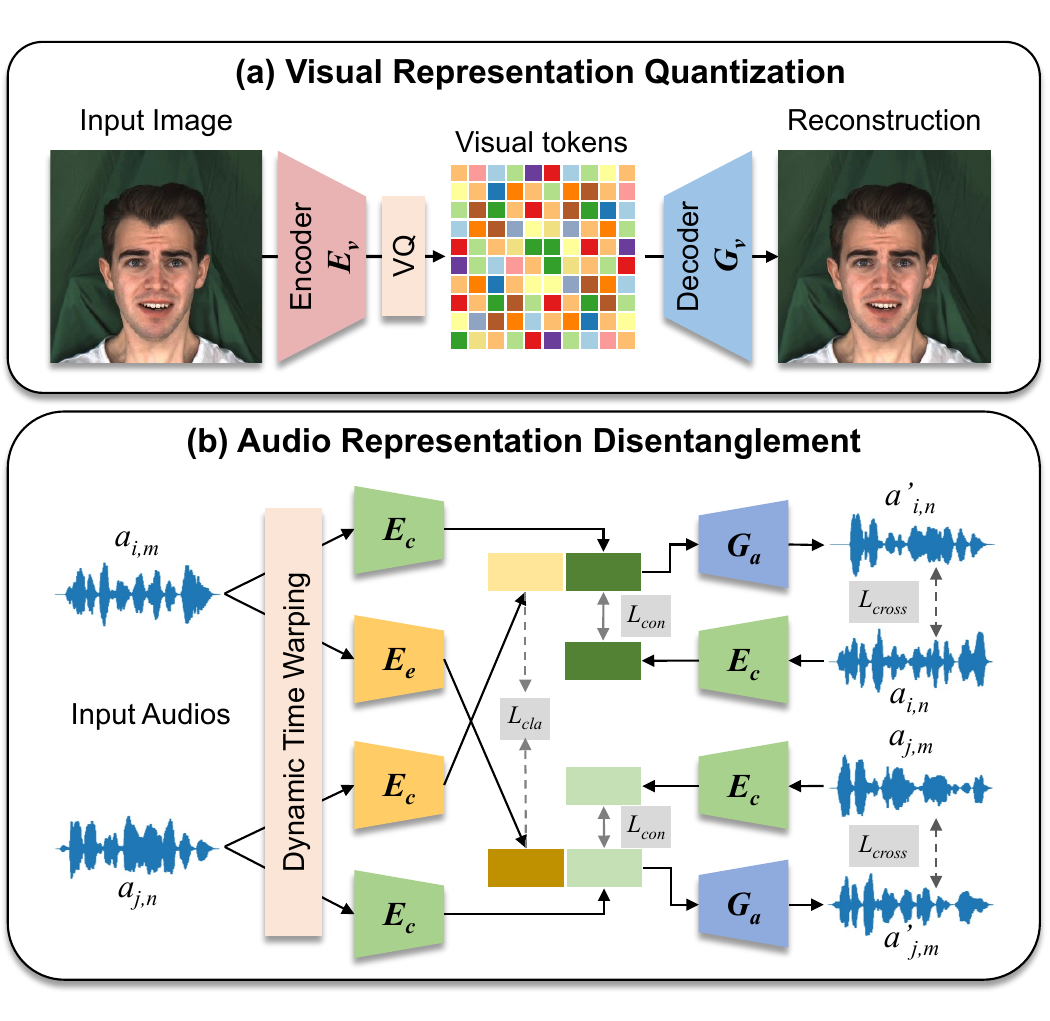}
	\caption{Illustration of the audio and visual pre-training strategies that mainly includes two parts: 
		(a) \textbf{Visual representation quantization.} A VQ-GAN-based generative network \cite{esser2021taming} that quantizes the facial images into visual tokens;
		(b) \textbf{Audio representation disentanglement.} A cross-reconstruction network  \cite{ji2021audio} that extracts disentangled emotion components and content components from audio signals.
	}
	\label{pretrain}
\end{figure}

In this paper, we design an autoregressive framework based on the discrete representation to tackle the emotion-controllable TFG task.
Unlike previous TFG methods \cite{chen2019hierarchical, chen2020talking, zhou2020makelttalk, ji2021audio, sinha2022emotion} that use landmark to represent the images,  a VQ-GAN-based  architecture \cite{esser2021taming} is leveraged to quantize the facial images into visual tokens.
The critical challenge is merging unique emotions into visual tokens to synthesize emotion-controllable facial videos.
To fulfill this goal, we first screen out the token sequences specific to the face region and propose an emotion-anchor (EA) representation to integrate the emotions and facial tokens through a local cross-attention operation.
The EA incorporates information about the emotional relevance of different facial areas.
Then we replace the visual tokens of input faces with the corresponding EA representations and further leverage this new sequence as conditions to generate new token indices sequences through  an autoregressive transformer.
The new token index sequences are leveraged to synthesize the final facial videos.
This latent transformer learns to bridge the gap between modalities and models the long-range interactions within these consituent  compositions.
Compared to traditional CNN-based generative methods that suffer from locality bias problem, this autoregressive manner generative network can make a more holistic understanding of the global composition of visual tokens, and achieve emotion-controllable generation.
At inference stage, emotion-controllable talking face generation can be achieved by integrating the emotional information extracted from the given audio into EA representations.
Extensive experiments on the MEAD dataset show that our method achieves significant improvements over the state-of-art approaches.

Our contributions are summarized as follows:
\begin{itemize}
	\item For the first time, we propose a discrete visual manipulation framework to tackle the audio-driven talking face generation task, which achieves emotion-controllable video synthesis. 
	\item To address the matching problem between audio components and discrete visual tokens, we elaborate a novel emotion-anchor (EA) representation to integrate the emotional information into facial tokens.
	\item A autoregressive transformer is introduced to capture the long-range interactions within the modalities and further generate emotion-controllable videos. Moreover, experiments  show that our method achieves significant improvements. 
\end{itemize}

\section{Related Work}

\subsection{Talking Face Generation} 
The audio-driven talking face generation has valuable applications in entertainment, attracting great interest in computer vision and computer graphics.
Earlier studies of talking face generation mainly focus on generating a specific identity from the dataset based on the given audio.
Chung \textit{et al.}  \cite{chung2017you} propose the first lip-synced video generation method in an image-to-image translation manner.
Subsequently, Kumar \textit{et al.} propose ObamaNet \cite{kumar2017obamanet} that uses time-delay LSTM to predict the representation of the mouth shape and further generates photo-realistic lip-sync videos.
Chen \textit{et al.} \cite{chen2018lip} propose a method that considers correlations among speech and lip movements while generating multiple lip images to synthesize talking faces of arbitrary identities.
In order to better extract related information from the speech, Zhou \textit{et al.} \cite{zhou2019talking} and Song \textit{et al.} \cite{song2018talking} leverage the disentangled audio-visual representation and recurrent neural networks, respectively.
Moreover, Chen \textit{et al.} \cite{chen2019hierarchical} design a hierarchical structure that leverages facial landmarks as intermediate representation and  generates talking faces based on the landmarks.

However, the above methods mainly focus on the accuracy of synchronization between generated mouths and audio, which ignore the expression or head movements.
To generate more talking head animation related to the identity, Zhou \textit{et al.} \cite{zhou2020makelttalk} propose a Speaker-Aware Animation module that maps the disentangled audio content to landmark displacements synchronizing the lip.
Song \textit{et al.} \cite{song2022everybody} propose a method that factorizes each target video frame into orthogonal parameter spaces and constructs a photo-realistic video via monocular 3D face reconstruction.
Recently Wang \textit{et al.} \cite{wang2020mead} released the MEAD dataset and proposed a method that renders emotional talking face videos by manipulating the upper and lower part of the face, respectively. 
Furthermore, Ji \textit{et al.} \cite{ji2021audio} propose a cross-reconstructed emotion disentanglement network to decompose the content and emotion of the audio and achieve varying emotional video portraits generation  by interpolating the emotion latent space.
Sinha \textit{et al.} propose an Emotion-Controllable Generalized (ECG) generation method \cite{sinha2022emotion} that can generalize to arbitrary faces.
In contrast to the previous landmark-based methods, our proposed method discretely manipulates facial emotions.

\begin{figure*}[tb]
	\centering
	\includegraphics[width=0.82\linewidth]{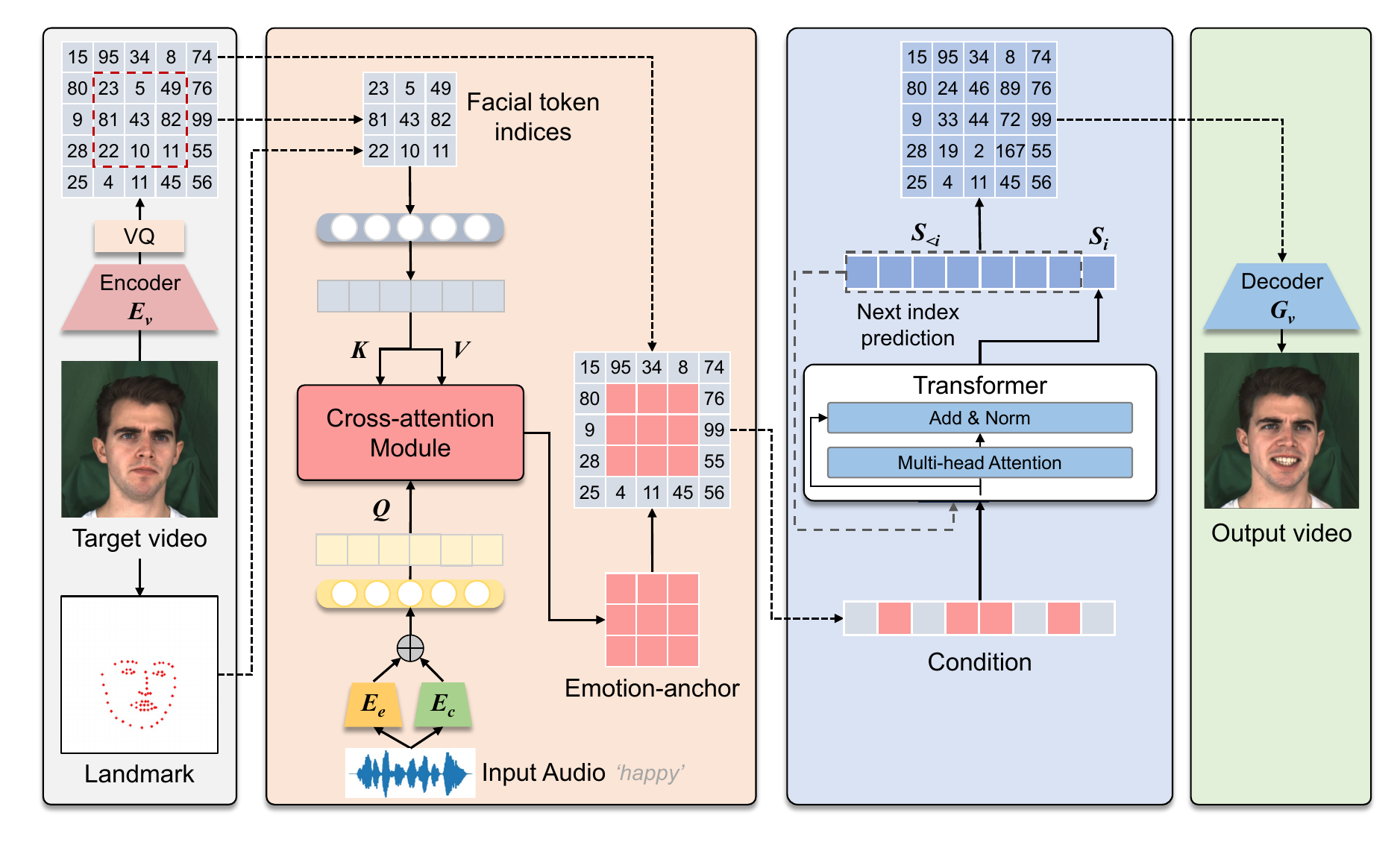}
	\caption{Overview of our proposed method.
		We use pre-trained encoders to obtain the sequence of the visual token index, the disentangled components from audio signals, and facial landmarks.
		Subsequently, we employ a cross-attention module to integrate emotional information into facial tokens.
		Then, we introduce an autoregressive transformer to predict the next index of visual tokens, given the condition  and previous indices.
		Finally, the manipulated video is generated by feeding the predicted sequence into the visual decoder.
}
	\label{framework}
\end{figure*}

\subsection{Conditional Emotion Generation} 
Conditional emotion generation \cite{jia2019gan} has long been studied in many research fields due to its importance for image generation.
Progress has been made in unsupervised image translation \cite{zhu2017unpaired, isola2017image, huang2018multimodal} for emotion-conditioned generation.
Pumarola \textit{et al.}  introduce an unsupervised framework named GANimation \cite{pumarola2018ganimation}, which is able to generate continuous facial expressions by representing facial emotions as action unit activation.
Ding \textit{et al.} design a novel encoder-decoder architecture \cite{ding2018exprgan} to control expression intensity continuously by learning an expressive and compact expression code. 
Vougioukas \textit{et al.} \cite{vougioukas2020realistic} design a specific discriminator that focuses on achieving realistic expression generation.
In this paper, we extract emotion components through the cross-reconstruction network and propose the emotion-anchor representation to integrate conditional emotion into visual tokens and perform emotion-controllable generation.

\section{Methodology}

\subsection{Framework Overview}
Our proposed emotion-controllable talking face generation method mainly contains two stages:
(1) The first stage is introduced and illustrated in Fig. \ref{pretrain}, which contains multimodal representation pre-training.
The audio representation network leverages a cross-reconstruction module \cite{ji2021audio} to disentangle the emotion and content from the audio. 
Meanwhile, the visual representation network adopts a VQ-GAN-based encoder-decoder \cite{esser2021taming} to learn the facial quantization representation;
(2) In the second stage, we design a novel emotion-anchor (EA) representation to integrate the emotions and facial tokens through a local cross-attention operation.
Then an autoregressive transformer \cite{vaswani2017attention} is leveraged to predict the distribution of possible next indices of visual tokens given the conditions, as shown in Fig. \ref{framework}. 
After index autoregression, the predicted sequence is fed into the pre-trained visual decoder to obtain the final edited videos.
Finally, we present the training objectives.

\subsection{Audio-Visual Data Preprocess}
\noindent\textbf{Audio Representation Disentanglement:}
The original audio signals contain the emotion and content components that are inherently coupled.
To achieve emotion conditional control for talking face generation, it is necessary to learn powerful emotion encoder and content encoder to extract independent representations.
Following \cite{ji2021audio}, we pre-train the cross-reconstruction network \cite{aberman2019learning} for audio emotion disentanglement, as shown in Fig. \ref{pretrain}(b).
Specifically, the input representations are the Mel Frequency Cepstral Coefficients (MFCC) \cite{logan2000mel} of the input audios.
Meanwhile, the encoders $E_c$ and $E_e$ to extract  content components and emotion components, respectively.
A decoder $G_a$ is used to reconstruct features.
The training of the audio cross-reconstruction network is supervised by four parts, including cross-reconstruction loss, self-reconstruction loss, classification loss, and content loss: 
 \begin{equation}
 	\begin{aligned}
 		\mathcal{L}_{\mathrm{aud}}(E_c, E_e, G_a) = \mathcal{L}_{\mathrm{cross}} + \mathcal{L}_{\mathrm{self}} +  \mathcal{L}_{\mathrm{cla}} + \mathcal{L}_{\mathrm{con}}.
 	\end{aligned}
 \end{equation}
After disentanglement pre-training, we freeze the parameters of emotion encoder $E_e$ and content encoder $E_c$.

\noindent\textbf{Face detection and landmark detection:} 
To better handle the facial emotional manipulation, we first perform face detection, cropping, and resizing them to 256 $\times$ 256 resolution by using an off-the-shelf facial detector  \cite{zhang2016joint}.
Then for each frame, we leverage FAN \cite{bulat2017far} to obtain the 68-point facial landmarks.
The detected landmarks are further leveraged to confirm the indices of the facial tokens.
Afterward, we estimate the faces' central coordinates $(x_c,y_c)$ based on the detected facial landmarks.

\noindent\textbf{Visual Representation Quantization:}
\label{sec:visual}
To better embed facial images into latent representations, we use a VQ-GAN-based encoder-decoder and pre-train it beforehand, as shown in Fig. \ref{pretrain}(a). 
The VQ-GAN model quantizes images into discrete visual tokens, primarily consisting of three components:
\begin{itemize}
	\item an image encoder $E_v$ that encodes images $x \in  \mathbb{R}^{H \times W \times 3}$ into embedding $e_x=E_v(x)$ with down-sampling ratio $n$, where $h=H/n, w=W/n$;
	\item a codebook $\mathcal{Z}=\{z_k\} \in \mathbb{R}^d, k \in 1,2,...K$ that contains $K$ visual tokens;
	\item a decoder $G_v$ with up-sampling ratio $n$ that reconstructs the image $\hat{x}$ from the visual tokens $z_k$.
\end{itemize}
Given a facial image $x$, the encoder $E_v$  embeds it  into a low-dimensional latent vector $e_x$.
Then $e_x$ is discretized by a subsequent element-wise quantization $\mathbf{q}(\cdot)$ to get both quantized index $z \in \mathbb{R}^{h \times w}$ and closet codebook entry $z_{\mathbf{q}} \in \mathbb{R}^{h \times w \times d}$.
The encoder-decoder and codebook are trained end-to-end via several loss functions:
\begin{equation}
	\begin{aligned}
		\mathcal{L}_{\mathrm{VQGAN}}(E_v,G_v) = \mathcal{L}_{\mathrm{VQ}} + \mathcal{L}_{\mathrm{adv}} + \mathcal{L}_{\mathrm{per}},
	\end{aligned}
\end{equation}
where $\mathcal{L}_{\mathrm{VQ}}$, $\mathcal{L}_{\mathrm{adv}}$, and $ \mathcal{L}_{\mathrm{per}}$ are VQ-VAE loss \cite{van2017neural}, adversarial loss, and perceptual loss \cite{johnson2016perceptual}, respectively.
The VQ-VAE loss is as follows:
\begin{equation}
	\begin{aligned}
		\mathcal{L}_{\mathrm{VQ}}(E_v,G_v,\mathcal{Z})&=  \Vert x - \hat{x} \Vert^2 + \Vert \mathrm{sg}[E_v(x)] - z_{\mathbf{q}}\Vert_2^2 \\
		& + \beta\Vert \mathrm{sg}[z_{\mathbf{q}}] - E_v(x)\Vert_2^2,
	\end{aligned}
\end{equation}
where $ \mathcal{L}_{\mathrm{rec}}=\Vert x - \hat{x} \Vert^2$ is the reconstruction loss, $\mathrm{sg}[\cdot]$ denotes the stop-gradient operation, and $\Vert \mathrm{sg}[z_{\mathbf{q}}] - E_v(x)\Vert_2^2$ is the commitment loss with weighting factor $\beta$.
After this pre-training stage, parameters in $E_v$ and $G_v$ are frozen.

With pre-trained encoders available, we use them to extract latent features for emotional autoregressive training, as shown in Fig. \ref{framework}.
More precisely, the quantized encoding of an image $x$ is given by $z_{\mathbf{q}}=\mathbf{q}(E_v(x)) \in \mathbb{R}^{h \times w \times d}$ and is equivalent to a sequence $s \in \{1,2,…,K\}^{h \times w}$ of indices from the codebook $\mathcal Z$, which is obtained by replacing each code by its index in the codebook $\mathcal Z$:
\begin{equation}
	s_{i,j} = k ~\mathrm{such~that}~ (z_{\mathbf{q}})_{i,j} = z_k,
\end{equation}
where $i$ and $j$ are the spatial coordinates of the image patches.
Thus, an image can be represented by a discrete sequence $s$ that contains indices from the codebook.
Meanwhile, by mapping the indices of $s$ back to their corresponding codebook entries $z_{\mathbf{q}}=(z_{s_{i,j}})$, we can recover the original image through  $\hat{x}=G_v({z_{\mathbf{q}}})$.

\begin{table*}[!t]
	\centering
	\small
	\setlength{\tabcolsep}{10pt}
	\renewcommand{\arraystretch}{1}
	\caption{The quantitative comparisons with SOTA methods.
		The evaluation metrics are leveraged to evalutate the  landmark accuracies and video qualities of the results by comparing them with the ground truth.}
	\begin{tabular}{c|c|c|c|c|c|c|c}
		\toprule
		Model/Metrics                           &M-LD$\downarrow$   &M-LVD$\downarrow$ &F-LD$\downarrow$  &F-LVD$\downarrow$ &SSIM$\uparrow$ &PSNR$\uparrow$ &FID$\downarrow$   \\		
		\midrule
		MEAD \cite{wang2020mead}     &2.52     &2.28   &3.16  &2.01  &0.68  &28.61  &22.52\\
		EVP \cite{ji2021audio}              &2.45     &1.78    &3.01  &1.56  &0.71  &29.53  &\textbf{7.99}\\
		ECG \cite{sinha2022emotion}    &2.18     &\textbf{0.77}    &1.24  &\textbf{0.50}  &0.77  &30.06  &35.41\\
		Ours                                             &\textbf{2.10}     &1.32    &\textbf{1.21} &0.98  &\textbf{0.78} &\textbf{30.78}&16.45\\
		\bottomrule
	\end{tabular}
	
	\label{tab:1}
\end{table*}

\subsection{Emotion-Anchor Representation}

In order to align the emotion and facial motions, we employ a cross-attention module to capture the local relationship between the emotion and the facial tokens, dubbed emotion-anchor (EA) representation.
For the audio, we use pre-trained audio emotion encoder $E_e$ and content encoder $E_c$ to get the corresponding representations.
Then, we add these two representations and perform a learnable encoder to compute the latent embedding $e_{e} \in \mathbb{R}^{d}$, where $d$ is the hidden size.
For the facial motions, a facial image can be represented by a token index sequence after quantization.
The emotional information is reflected only by the tokens associated with the face, regardless of the background.
Thus, we first calculate the central coordinates $(x_c,y_c)$ to estimate the facial regions $r_f=\{(i,j) | {i\in [ [x_c]-{\mathtt x},  [x_c]+{\mathtt x} ],  j \in [ [y_c]-{\mathtt y},  [y_c]+{\mathtt y}}\}$, and the facial token indices $s_f= \{ s_{i,j} |  (i,j)\in r_f\}$, where $[\cdot]$ is the round operation and $\mathtt x/\mathtt y$ are both hyperparameters.
Then, the facial indices sequence $s_f$ is converted into facial token embedding $e_f \in \mathbb{R}^{[(2\mathtt x +1)\cdot (2\mathtt y +1)]\times d}$ by another learnable encoder.
The emotion embedding $e_e$ is used as a query, and the facial token embedding $e_f$ is used as both key and value of the cross-attention module.
The cross-attention module can model the distribution of facial tokens for different emotions and aggregate the local  motion information:
\begin{equation}
	\begin{split}
	& Q=e_eW^e, K=e_fW^k, V=e_fW^v, \\
	& Attn = \mathrm{MultiHead}(Q,K,V),\\
	& \mathrm{EA} =\mathrm{FFN}(e_e,Attn),
	\end{split}
\end{equation}
where  $\mathrm{MultiHead}$ and $\mathrm{FFN}$ represent multi-head attention and feed-forward network, respectively.
Therefore, the emotion-anchor representation $\mathrm{EA}\in \mathbb{R}^{[(2\mathtt x +1)\cdot (2\mathtt y +1)]\times d}$ stores the emotion and facial motion information.

\begin{figure*}[tb]
	\centering
	\includegraphics[width=0.65\linewidth]{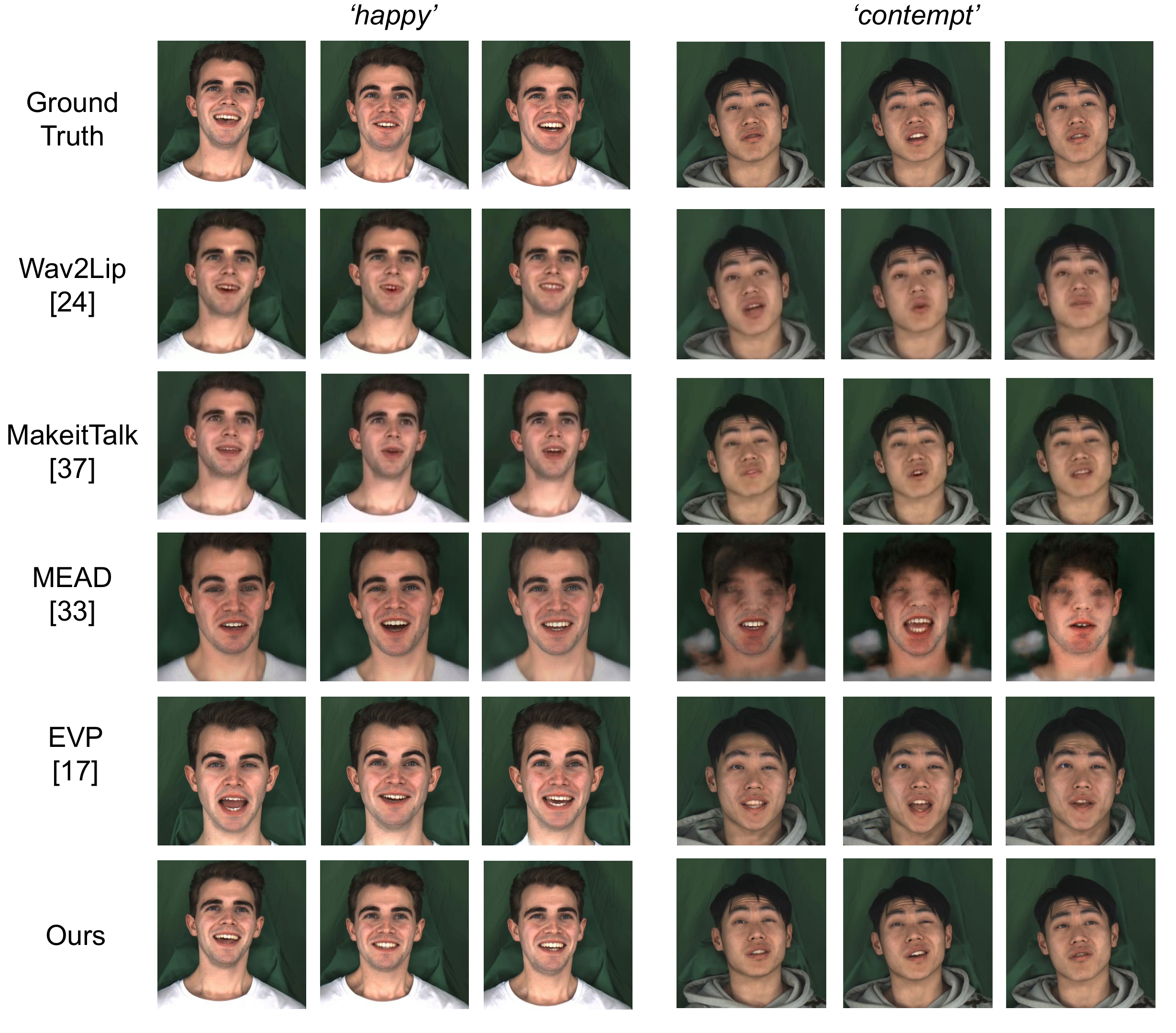}
	\caption{The qualitative results of our method with several SOTA talking face generation methods on the MEAD dataset.
		The experiments are conducted on two examples with different speech content and emotions.
		Note that we use the first frame of the target video for MakeitTalk \cite{zhou2020makelttalk}, Wav2Lip \cite{prajwal2020lip}, and MEAD \cite{wang2020mead} as they edit a target image rather than a target video.
	}
	\label{ex11}
\end{figure*}

\subsection{Emotional Fusion Autoregression}

Following \cite{esser2021taming}, we treat the facial image generation process as an autoregressive learning, \textit{i.e.} next-index prediction.
We use a transformer decoder model as an autoregressive learner. 
Given indices $s_{<i}$, the latent transformer learns to predict the distribution of the next indices and computes the likelihood of the complete representation as $p(s)=\prod_{i}p(s_i|s_{<i})$.
Furthermore, we use the emotion-anchor representations as additional control conditions to achieve emotional autoregressive training.
In order to fusion the information of EA, we replace the indices in the sequence $s$ with the characteristic of the corresponding position in EA as the condition for image generation, as shown in Fig. \ref{framework}:
\begin{eqnarray}
	\begin{split}
	c_{i,j}=
	\begin{cases}
		s_f & \mathrm{if} \ (i,j) \in r_f \\
		s &  \mathrm{else}
	\end{cases}
\end{split},
\end{eqnarray}
Then we use this new sequence as conditions for conditional autoregressive training: 
\begin{equation}
	p(s|c)=\prod_{i}p(s_i|s_{<i}, c),
\end{equation}

\subsection{The Training Objective}
Our proposed method contains the multimodal representation pre-training stage and the autoregressive training stage.
In the first stage, we pre-train the visual representation network and audio representation separately.
The training objective of the first stage is:
\begin{equation}
	\mathcal{L_{\mathrm{fir}}}=\mathcal{L}_{\mathrm{aud}}(E_e, E_c, G_a) + \mathcal{L}_{\mathrm{VQGAN}}(E_v,G_v).
\end{equation}

In the second stage, we train a cross-attention module and an autoregressive transformer that learns to predict the index sequence of visual tokens driven by the given conditions.
Due to the problem of unpaired data, we use a self-reconstruction strategy to train the autoregressive prediction by directly maximizing the log-likelihood of the data representations:
\begin{equation}
	\mathcal{L_{\mathrm{auto}}}=\mathbb{E}_{x \sim p(x)}[-\log p(s|c)].
\end{equation}
To further enhance the continuity between frames, we introduce additional continuity loss to constrain the degree of difference between the index sequence of the predicted image and the previous frame:
\begin{equation}
	\mathcal{L_{\mathrm{conti}}}=\mathbb{E}_{x_{pre} \sim p(x_{pre})}[-\log p(s|c)],
\end{equation}
where $x_{pre}$ is the previous frame image of $x$.
Thus, the training objective of the second stage is:
\begin{equation}
	\mathcal{L_{\mathrm{sec}}}=\mathcal{L_{\mathrm{auto}}} + \lambda\mathcal{L_{\mathrm{conti}}}.
\end{equation}
We set $\lambda=0.5$ empirically during training.

\section{Experiments}

In this section, we first introduce the involved experimental dataset with related data pre-processing methods.
Next, we give the implementation details.
Then, we present our proposed method's quantitative results and qualitative evaluation.
Finally, the ablation studies will be presented to prove the effectiveness of our method.

\subsection{Datasets}

We use the emotional audio-visual dataset MEAD \cite{wang2020mead} for talking face generation experiments.
The MEAD dataset contains high-quality talking face videos of 60 actors/actresses with eight  emotions and multiple speech audios.
We choose three actors and split them into train/test sets for training and testing, respectively.
For visual data, all the emotional talking face videos are converted to 25 fps.
For the audio stream, the speech sample rate is set to be 16kHz.
Following the design in \cite{chen2019hierarchical}, we extract the 28$\times$12 MFCC feature corresponding to each frame in the video.

\subsection{Implementation Details}

In the first pre-training stage, we use the audio emotion encoder  pre-trained through an emotion classification task \cite{ooi2014new} to initialize our emotion encoder $E_e$.
Moreover, the content encoder $E_c$ is pre-trained on LRW \cite{chung2016lip}, a lip-reading dataset with barely any emotion. 
Meanwhile, we set the codebook size $|\mathcal{Z}|=K=2048$, the down-sampling ratio $n=16$, and $d=256$ in visual representation pre-training.
In other words, the length of the predicted index sequence $s=16\times16=256$.

Next, in the second stage, we leverage the well-trained multimodal encoder-decoders and freeze their parameters to extract the discrete representation, respectively.
In the emotional autoregressive training, we use a GPT2-medium architecture \cite{radford2019language} as our index sequence predictor.
We set $\mathtt{x}=\mathtt{y}=5$ empirically.
We train the second stage on NVIDIA TITAN RTX GPUs  using Adam Optimizer, with a learning rate of 4.5e-06 and a batch size of 16.

\subsection{Quantitative Results}
For quantitative evaluation, we compare our proposed method with  TFG methods MEAD \cite{wang2020mead}, EVP \cite{ji2021audio}, and  ECG \cite{sinha2022emotion} that are trained and tested on the MEAD dataset.
We evaluate the generated animation results in three essential aspects: landmark validation (M-LD, M-LVD, F-LD, F-LVD), video quality (SSIM, PSNR, FID), and user studies.
The quantitative results are presented in Tab. \ref{tab:1}.

\begin{figure*}[t]
	\centering
	\includegraphics[width=0.78\linewidth]{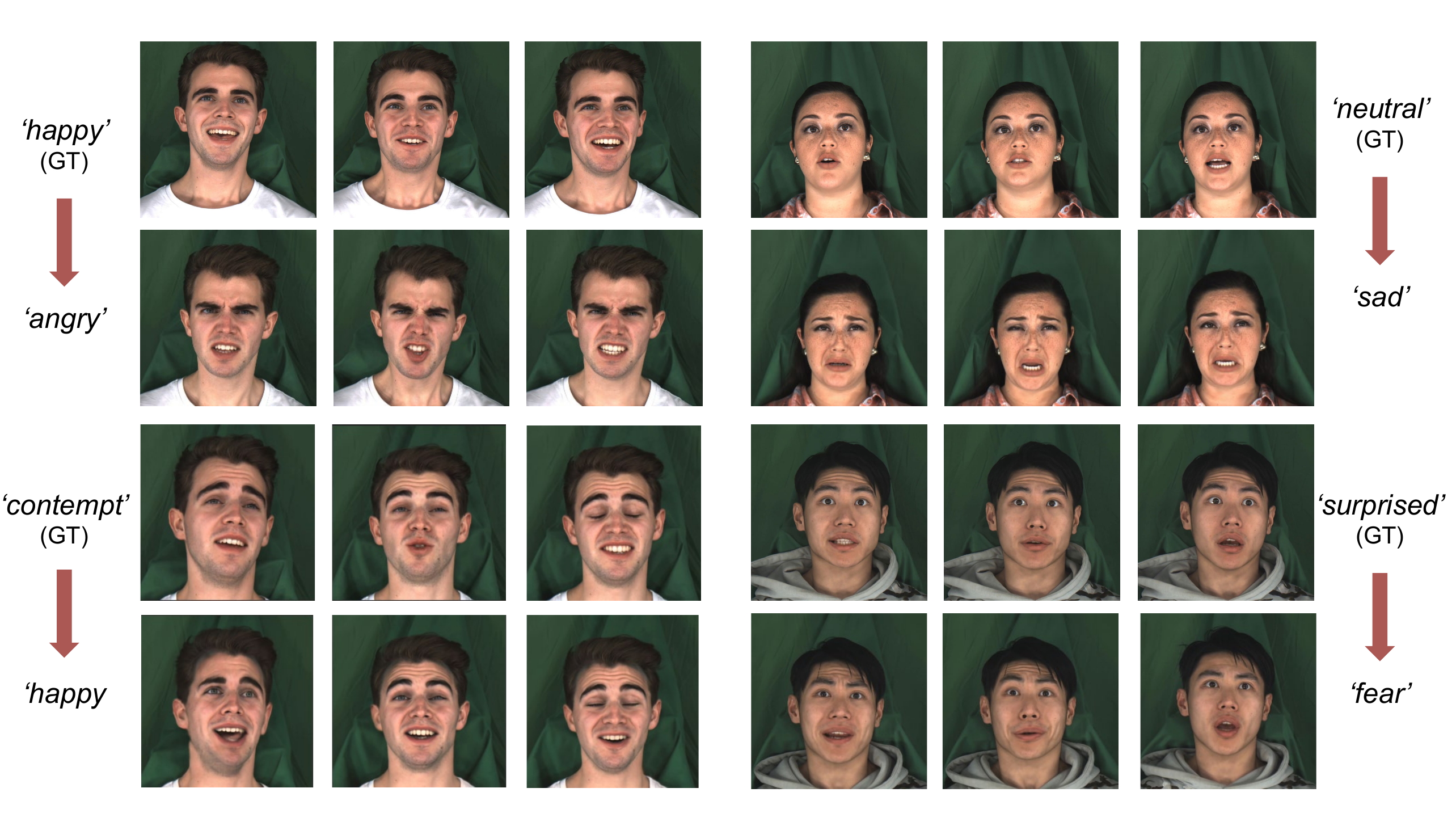}
	\caption{Illustration of the qualitative results with different emotions and identities.
		Due to discrete representation, we can achieve emotion-controllable talking face generation.
		The two ends of the arrow represent the initial emotion and the target emotion, respectively.
	}
	\label{ex22}
\end{figure*}

\begin{figure}[tb]
	\centering
	\includegraphics[width=1\linewidth]{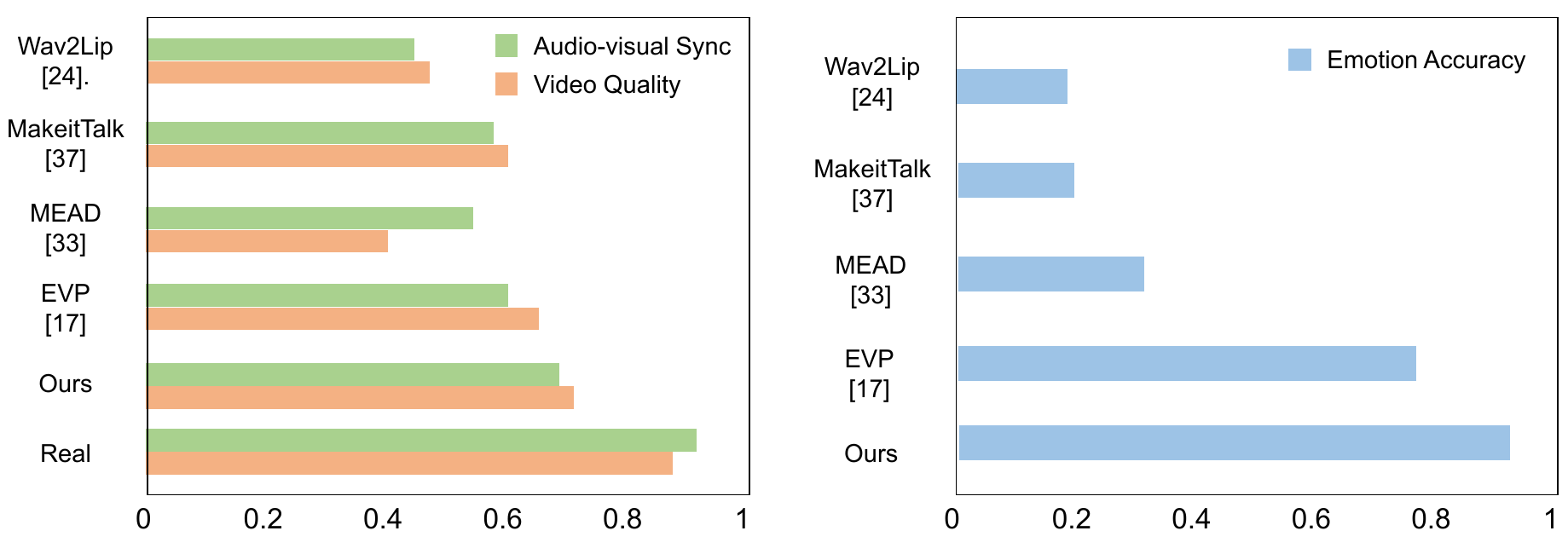}
	\caption{User study results of audio-visual synchronization,
		video quality and emotion accuracy.
	}
	\label{user}
\end{figure}

\subsubsection{Landmark Validation}
To quantitatively evaluate different methods, we extract facial landmarks from the generated results and the ground truth for comparison.
Following \cite{ji2021audio}, the metrics of Landmark Distance (LD) and Landmark Velocity Difference (LVD) are utilized to evaluate the accuracy of facial motions. 
LD represents the average Euclidean distance between generated and recorded landmarks, and velocity means the difference in landmark locations between consecutive frames.
So LVD represents the average velocity differences of landmark motions between two sequences.
We adopt LD and LVD on the mouth and face area to evaluate how well the synthesized video represents accurate lip movements and facial expressions separately. 
Thanks to the discrete representation, our proposed method is capable of modeling the relationship between the audio representations and the compositions of images.
Due to the weak constraints of the discrete representaion on the spatial location, some results could make better on LVD. 

\subsubsection{Video Quality}
To evaluate the quality of the generated images of different methods, we use the SSIM \cite{wang2004image}, PSNR, and FID \cite{heusel2017gans} scores.
Our method outperforms the SOTA methods in most of the texture quality metrics.
Due to that the image generator 
EVP outperforms all the methods in FID because they train person-specific texture models.

\begin{table}[!t]
	\centering
	\small
	\renewcommand{\arraystretch}{1}
	\caption{Quantitative ablation result of landmarks for our proposed method with different losses and facial regions.}
	\begin{tabular}{c|c|c|c|c}
		\toprule
		Method/Score                           &M-LD$\downarrow$   &M-LVD$\downarrow$ &F-LD$\downarrow$  &F-LVD$\downarrow$   \\		
		\midrule
        Ours w/o $\mathcal{L_{\mathrm{conti}}}$   &2.48     &1.89   &1.65  &1.37 \\
		Ours($\mathtt{x}$=$\mathtt{y}$=3)        &2.30     &1.67    &1.33  &1.16\\
		Ours($\mathtt{x}$=$\mathtt{y}$=7)          &2.11    &1.31    &1.22   &0.98\\
		Ours        &2.10    &1.32    &1.21   &0.98\\
		\bottomrule
	\end{tabular}
	
	\label{tab:2}
\end{table}

\subsubsection{User Studies}

To quantify the quality (including the accuracy of emotion and facial motion) of the synthesized video clips, we conducted several user studies on the MEAD dataset with MakeitTalk \cite{zhou2020makelttalk}, Wav2Lip \cite{prajwal2020lip}, MEAD \cite{wang2020mead}, and EVP \cite{ji2021audio}.
We generate three video clips for each of the eight emotion categories and each of three actors.
Then we ask 25 participants to rate the total videos generated from our method and other methods.
They are evaluated w.r.t three different criteria: whether the synthesized talking face video is realistic, whether the face motion sync with the speech, and the accuracy of the generated facial emotion. 
We first ask the participants to judge the given video upon audio-visual synchronization and video quality and score from 1 (worst) to 5 (best).
Then, they are asked to choose the emotion category for the generated video without voice.
We finished these questionnaires and showed the results in Fig. \ref{user}. 
As can be seen, our method obtains the highest score on visual quality and audio-visual sync apart from the real data. 
We also achieve the highest accuracy on emotion classification compared with other methods.

\begin{figure}[tb]
	\centering
	\includegraphics[width=0.85\linewidth]{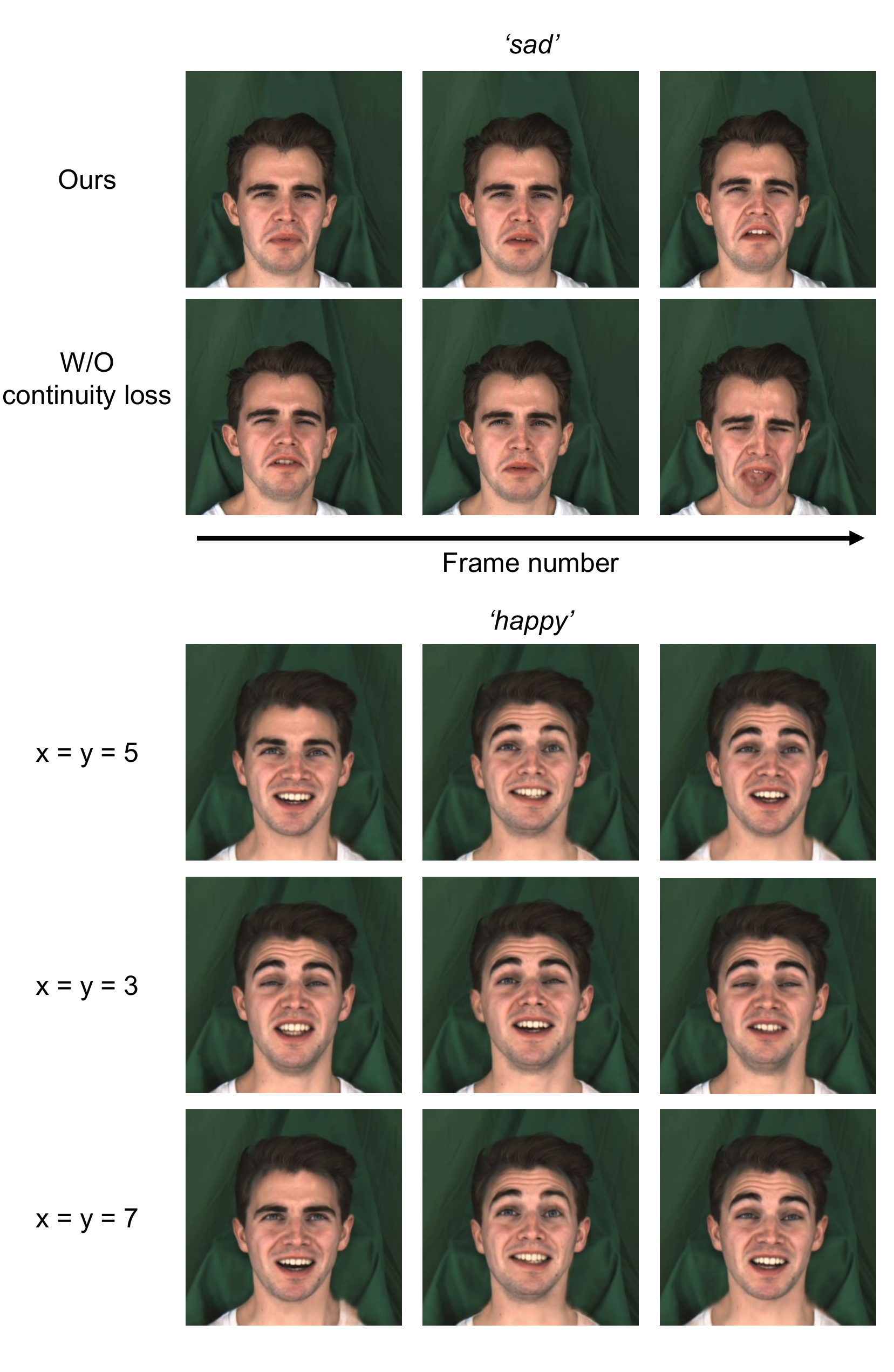}
	\caption{Qualitative ablation for our proposed method.
		The top part is a series of chronological frames that w/o continuity loss.
		The bottom part is the ablation experiment of setting the size of the facial region.
		The comparison results clearly show that the proposed continuity loss and the choice of $\mathtt{x/y}$ are significant to the synthesized videos.	
	}
	\label{ab}
\end{figure}

\subsection{Qualitative Results}
For qualitative evaluation, we compare our proposed method with SOTA talking face generation methods: MakeitTalk \cite{zhou2020makelttalk}, Wav2Lip \cite{prajwal2020lip}, MEAD \cite{wang2020mead}, and EVP \cite{ji2021audio}.
MEAD and EVP are the most relevant works since they render emotion, while MakeitTalk and Wav2Lip are trained on other datasets.
We have evaluated all the methods using their publicly available pre-trained model.
The final animation results are illustrated in Fig. \ref{ex11}.
Due to the limitation of the dataset, MakeItTalk and Wav2Lip mainly complete lip-sync, rather than  render emotion onto faces. 
Since the publicly available pre-trained model for MEAD  is only trained for Subject 1 (Fig. \ref{ex11} left), their method is unable to generalize to other identities (Fig. \ref{ex11} right). 
EVP mainly renders the local facial expression, leading to mechanized expression of emotion.
Our method learns the global information between modalities and can produce better emotion and preserve identity.

We also conduct corresponding experiments on the proposed discrete representation for emotional TFG, as shown in Fig. \ref{ex22}.
Given the desired emotion, we can render this emotion onto a target video by extracting the corresponding emotion tokens in the control conditions.
Subsequently, by feeding the conditions into the well-trained transformers, we can obtain the indices sequence of the edited video.
Through the generator $G_v$, we can perform emotional talking face generation.
Note that the modified video keeps the same audio content.


\subsection{Ablation Study}

For emotion-controllable TFG, we conduct two ablation studies to evaluate the model structures' effectiveness.
Quantitative and qualitative results are shown in Tab. \ref{tab:2} and Fig. \ref{ab} that prove the contributions of each component., respectively.
The input conditions for transformer have no temporal supervisory information, which may break the relationship between frames.
We made ablation studies to validate the ability of continuity loss $\mathcal{L_{\mathrm{conti}}}$.
Our proposed continuity loss helps the model to capture the association information between frames.

Moreover, we also conduct experiments on the size of the facial region, i.e., the setting of hyperparameters $\mathtt{x}$ and $\mathtt{y}$.
It is obvious that the emotion-anchor representations could integrate more facial information by increasing the size of the facial region ($\mathtt{x/y}$ from 3 to 5), making the resulting face more emotionally responsive.
However, the generated results mostly stay the same but increase the cost of computation when increasing the size of face region  ($\mathtt{x/y}$ from 5 to 7).
This is because the irrelevant background information does not contain the relevant information of emotion after enlarging the region.

\section{Conclusion}

In this paper, we propose an autoregressive method based on discrete representation to tackle the emotion-controllable talking face generation task. 
Specifically, we first employ several pre-training strategies to disentangle the emotion and content components from audio signals and quantize videos into  visual tokens.
To better integrate audio emotion into visual token representations, we propose the emotion-anchor (EA) representation to fuse cross-modal information through a cross-attention module.
Subsequently, we introduce an autoregressive transformer to model the relationship between the global composition of visual tokens and the given condition.
By leveraging this latent transformer to capture the long-range interactions contained in quantized tokens, our proposed method can precisely achieve emotional generation  and improve the synthesized videos' authenticity.
Experiments show that our method achieves impressive performance on the MEAD dataset.

\clearpage
{\small
\bibliographystyle{ieee_fullname}
\bibliography{CVPR_ZZZ}

\begin{thebibliography}{10}\itemsep=-1pt

\bibitem{aberman2019learning}
Kfir Aberman, Rundi Wu, Dani Lischinski, Baoquan Chen, and Daniel Cohen-Or.
\newblock Learning character-agnostic motion for motion retargeting in 2d.
\newblock {\em arXiv preprint arXiv:1905.01680}, 2019.

\bibitem{adalgeirsson2010mebot}
Sigurdur~Orn Adalgeirsson and Cynthia Breazeal.
\newblock Mebot: A robotic platform for socially embodied telepresence.
\newblock In {\em 2010 5th ACM/IEEE International Conference on Human-Robot
  Interaction (HRI)}, pages 15--22. IEEE, 2010.

\bibitem{bulat2017far}
Adrian Bulat and Georgios Tzimiropoulos.
\newblock How far are we from solving the 2d \& 3d face alignment problem?(and
  a dataset of 230,000 3d facial landmarks).
\newblock In {\em Proceedings of the IEEE International Conference on Computer
  Vision}, pages 1021--1030, 2017.

\bibitem{chen2020talking}
Lele Chen, Guofeng Cui, Celong Liu, Zhong Li, Ziyi Kou, Yi Xu, and Chenliang
  Xu.
\newblock Talking-head generation with rhythmic head motion.
\newblock In {\em European Conference on Computer Vision}, pages 35--51.
  Springer, 2020.

\bibitem{chen2018lip}
Lele Chen, Zhiheng Li, Ross~K Maddox, Zhiyao Duan, and Chenliang Xu.
\newblock Lip movements generation at a glance.
\newblock In {\em Proceedings of the European Conference on Computer Vision
  (ECCV)}, pages 520--535, 2018.

\bibitem{chen2019hierarchical}
Lele Chen, Ross~K Maddox, Zhiyao Duan, and Chenliang Xu.
\newblock Hierarchical cross-modal talking face generation with dynamic
  pixel-wise loss.
\newblock In {\em Proceedings of the IEEE/CVF conference on computer vision and
  pattern recognition}, pages 7832--7841, 2019.

\bibitem{chung2017you}
Joon~Son Chung, Amir Jamaludin, and Andrew Zisserman.
\newblock You said that?
\newblock {\em arXiv preprint arXiv:1705.02966}, 2017.

\bibitem{chung2016lip}
Joon~Son Chung and Andrew Zisserman.
\newblock Lip reading in the wild.
\newblock In {\em Asian conference on computer vision}, pages 87--103.
  Springer, 2016.

\bibitem{ding2018exprgan}
Hui Ding, Kumar Sricharan, and Rama Chellappa.
\newblock Exprgan: Facial expression editing with controllable expression
  intensity.
\newblock In {\em Proceedings of the AAAI conference on artificial
  intelligence}, volume~32, 2018.

\bibitem{edwards2016jali}
Pif Edwards, Chris Landreth, Eugene Fiume, and Karan Singh.
\newblock Jali: an animator-centric viseme model for expressive lip
  synchronization.
\newblock {\em ACM Transactions on graphics (TOG)}, 35(4):1--11, 2016.

\bibitem{eskimez2021speech}
Sefik~Emre Eskimez, You Zhang, and Zhiyao Duan.
\newblock Speech driven talking face generation from a single image and an
  emotion condition.
\newblock {\em IEEE Transactions on Multimedia}, 2021.

\bibitem{esser2021taming}
Patrick Esser, Robin Rombach, and Bjorn Ommer.
\newblock Taming transformers for high-resolution image synthesis.
\newblock In {\em Proceedings of the IEEE/CVF conference on computer vision and
  pattern recognition}, pages 12873--12883, 2021.

\bibitem{heusel2017gans}
Martin Heusel, Hubert Ramsauer, Thomas Unterthiner, Bernhard Nessler, and Sepp
  Hochreiter.
\newblock Gans trained by a two time-scale update rule converge to a local nash
  equilibrium.
\newblock {\em Advances in neural information processing systems}, 30, 2017.

\bibitem{huang2018multimodal}
Xun Huang, Ming-Yu Liu, Serge Belongie, and Jan Kautz.
\newblock Multimodal unsupervised image-to-image translation.
\newblock In {\em Proceedings of the European conference on computer vision
  (ECCV)}, pages 172--189, 2018.

\bibitem{isola2017image}
Phillip Isola, Jun-Yan Zhu, Tinghui Zhou, and Alexei~A Efros.
\newblock Image-to-image translation with conditional adversarial networks.
\newblock In {\em Proceedings of the IEEE conference on computer vision and
  pattern recognition}, pages 1125--1134, 2017.

\bibitem{ji2021audio}
Xinya Ji, Hang Zhou, Kaisiyuan Wang, Wayne Wu, Chen~Change Loy, Xun Cao, and
  Feng Xu.
\newblock Audio-driven emotional video portraits.
\newblock In {\em Proceedings of the IEEE/CVF conference on computer vision and
  pattern recognition}, pages 14080--14089, 2021.

\bibitem{jia2019gan}
Xiaoqi Jia, Jianwei Tai, Hang Zhou, Yakai Li, Weijuan Zhang, Haichao Du, and
  Qingjia Huang.
\newblock Et-gan: Cross-language emotion transfer based on cycle-consistent
  generative adversarial networks.
\newblock {\em arXiv preprint arXiv:1905.11173}, 2019.

\bibitem{johnson2016perceptual}
Justin Johnson, Alexandre Alahi, and Li Fei-Fei.
\newblock Perceptual losses for real-time style transfer and super-resolution.
\newblock In {\em European conference on computer vision}, pages 694--711.
  Springer, 2016.

\bibitem{kumar2017obamanet}
Rithesh Kumar, Jose Sotelo, Kundan Kumar, Alexandre de Br{\'e}bisson, and
  Yoshua Bengio.
\newblock Obamanet: Photo-realistic lip-sync from text.
\newblock {\em arXiv preprint arXiv:1801.01442}, 2017.

\bibitem{liu2015video}
Yilong Liu, Feng Xu, Jinxiang Chai, Xin Tong, Lijuan Wang, and Qiang Huo.
\newblock Video-audio driven real-time facial animation.
\newblock {\em ACM Transactions on Graphics (TOG)}, 34(6):1--10, 2015.

\bibitem{logan2000mel}
Beth Logan.
\newblock Mel frequency cepstral coefficients for music modeling.
\newblock In {\em In International Symposium on Music Information Retrieval}.
  Citeseer, 2000.

\bibitem{ooi2014new}
Chien~Shing Ooi, Kah~Phooi Seng, Li-Minn Ang, and Li~Wern Chew.
\newblock A new approach of audio emotion recognition.
\newblock {\em Expert systems with applications}, 41(13):5858--5869, 2014.

\bibitem{prajwal2020lip}
KR Prajwal, Rudrabha Mukhopadhyay, Vinay~P Namboodiri, and CV Jawahar.
\newblock A lip sync expert is all you need for speech to lip generation in the
  wild.
\newblock In {\em Proceedings of the 28th ACM International Conference on
  Multimedia}, pages 484--492, 2020.

\bibitem{pumarola2018ganimation}
Albert Pumarola, Antonio Agudo, Aleix~M Martinez, Alberto Sanfeliu, and
  Francesc Moreno-Noguer.
\newblock Ganimation: Anatomically-aware facial animation from a single image.
\newblock In {\em Proceedings of the European conference on computer vision
  (ECCV)}, pages 818--833, 2018.

\bibitem{radford2019language}
Alec Radford, Jeffrey Wu, Rewon Child, David Luan, Dario Amodei, Ilya
  Sutskever, et~al.
\newblock Language models are unsupervised multitask learners.
\newblock {\em OpenAI blog}, 1(8):9, 2019.

\bibitem{sinha2022emotion}
Sanjana Sinha, Sandika Biswas, Ravindra Yadav, and Brojeshwar Bhowmick.
\newblock Emotion-controllable generalized talking face generation.
\newblock {\em arXiv preprint arXiv:2205.01155}, 2022.

\bibitem{song2022everybody}
Linsen Song, Wayne Wu, Chen Qian, Ran He, and Chen~Change Loy.
\newblock Everybody’s talkin’: Let me talk as you want.
\newblock {\em IEEE Transactions on Information Forensics and Security},
  17:585--598, 2022.

\bibitem{song2018talking}
Yang Song, Jingwen Zhu, Dawei Li, Xiaolong Wang, and Hairong Qi.
\newblock Talking face generation by conditional recurrent adversarial network.
\newblock {\em arXiv preprint arXiv:1804.04786}, 2018.

\bibitem{van2017neural}
Aaron Van Den~Oord, Oriol Vinyals, et~al.
\newblock Neural discrete representation learning.
\newblock {\em Advances in neural information processing systems}, 30, 2017.

\bibitem{vaswani2017attention}
Ashish Vaswani, Noam Shazeer, Niki Parmar, Jakob Uszkoreit, Llion Jones,
  Aidan~N Gomez, {\L}ukasz Kaiser, and Illia Polosukhin.
\newblock Attention is all you need.
\newblock {\em Advances in neural information processing systems}, 30, 2017.

\bibitem{vougioukas2020realistic}
Konstantinos Vougioukas, Stavros Petridis, and Maja Pantic.
\newblock Realistic speech-driven facial animation with gans.
\newblock {\em International Journal of Computer Vision}, 128(5):1398--1413,
  2020.

\bibitem{wang2020mead}
Kaisiyuan Wang, Qianyi Wu, Linsen Song, Zhuoqian Yang, Wayne Wu, Chen Qian, Ran
  He, Yu Qiao, and Chen~Change Loy.
\newblock Mead: A large-scale audio-visual dataset for emotional talking-face
  generation.
\newblock In {\em European Conference on Computer Vision}, pages 700--717.
  Springer, 2020.

\bibitem{wang2004image}
Zhou Wang, Alan~C Bovik, Hamid~R Sheikh, and Eero~P Simoncelli.
\newblock Image quality assessment: from error visibility to structural
  similarity.
\newblock {\em IEEE transactions on image processing}, 13(4):600--612, 2004.

\bibitem{zhang2016joint}
Kaipeng Zhang, Zhanpeng Zhang, Zhifeng Li, and Yu Qiao.
\newblock Joint face detection and alignment using multitask cascaded
  convolutional networks.
\newblock {\em IEEE signal processing letters}, 23(10):1499--1503, 2016.

\bibitem{zhou2019talking}
Hang Zhou, Yu Liu, Ziwei Liu, Ping Luo, and Xiaogang Wang.
\newblock Talking face generation by adversarially disentangled audio-visual
  representation.
\newblock In {\em Proceedings of the AAAI conference on artificial
  intelligence}, volume~33, pages 9299--9306, 2019.

\bibitem{zhou2020makelttalk}
Yang Zhou, Xintong Han, Eli Shechtman, Jose Echevarria, Evangelos Kalogerakis,
  and Dingzeyu Li.
\newblock Makelttalk: speaker-aware talking-head animation.
\newblock {\em ACM Transactions on Graphics (TOG)}, 39(6):1--15, 2020.

\bibitem{zhu2017unpaired}
Jun-Yan Zhu, Taesung Park, Phillip Isola, and Alexei~A Efros.
\newblock Unpaired image-to-image translation using cycle-consistent
  adversarial networks.
\newblock In {\em Proceedings of the IEEE international conference on computer
  vision}, pages 2223--2232, 2017.

\end{thebibliography}
}

\end{document}